\UseRawInputEncoding
\documentclass[conference]{IEEEtran}
\IEEEoverridecommandlockouts
\usepackage{cite}
\usepackage{amsmath,amssymb,amsfonts}
\usepackage{algorithmic}
\usepackage{graphicx}
\usepackage{textcomp}
\usepackage{hyperref}
\usepackage{xcolor}
\usepackage{float}
\def\BibTeX{{\rm B\kern-.05em{\sc i\kern-.025em b}\kern-.08em
    T\kern-.1667em\lower.7ex\hbox{E}\kern-.125emX}}
\begin{document}

\title{Terrain Adaptive Gait Transitioning for a Quadruped Robot using Model Predictive Control\\}

\author{\IEEEauthorblockN{Prathamesh Saraf$^{1}$, Abhishek Sarkar$^{2}$, Arshad Javed$^{2}$}
\IEEEauthorblockA{\textit{$^{1}$Department of Electrical and Electronics Engineering} \\
\textit{$^{2}$Department of Mechanical Engineering} \\
\textit{Birla Institute of Technology and Science, Pilani - Hyderabad, India}\\
pratha1999@gmail.com, abhisheks@hyderabad.bits-pilani.ac.in, arshad@hyderabad.bits-pilani.ac.in}}
\maketitle

\begin{abstract}
 Legged robots can traverse challenging terrain, use perception to plan their safe foothold positions, and navigate the environment. Such unique mobility capabilities make these platforms a perfect candidate for scenarios such as search and rescue, inspection, and exploration tasks. While traversing through such terrains, the robot's instability is a significant concern. Many times the robot needs to switch gaits depending on its environment. Due to the complex dynamics of quadruped robots, classical PID control fails to provide high stability. Thus, there is a need for advanced control methods like the Model Predictive Control (MPC) which uses the system model and the nature of the terrain in order to predict the stable body pose of the robot. The controller also provides correction to any external disturbances that result in a change in the desired behavior of the robot. The MPC controller is designed in MATLAB, for full body torque control. The controller performance was verified on Boston Dynamics Spot in Webots simulator. The robot is able to provide correction for external perturbations up to 150 N and also resist falls till 80 cm.
\end{abstract}


\section{Introduction}
Legged robots are gaining tremendous popularity due to their all-terrain locomotion properties. The advantage of legged robots is high-speed locomotion along with less surface coverage which makes them suitable for many applications like rescue, inspection, and exploration. Among all legged robots, quadruped robots are the most researched as they provide maximum functionality with a less complex control. Quadruped robots are designed with three main functionalities; high-speed locomotion~\cite{MIT}, high stability for all kinds of terrains~\cite{ANYmal}~\cite{lit1}, and high jumping capabilities. This work mainly focuses on the stability of quadruped robots on challenging terrains.\\
\begin{figure}[htbp]
    \centerline{\includegraphics[width = 244pt]{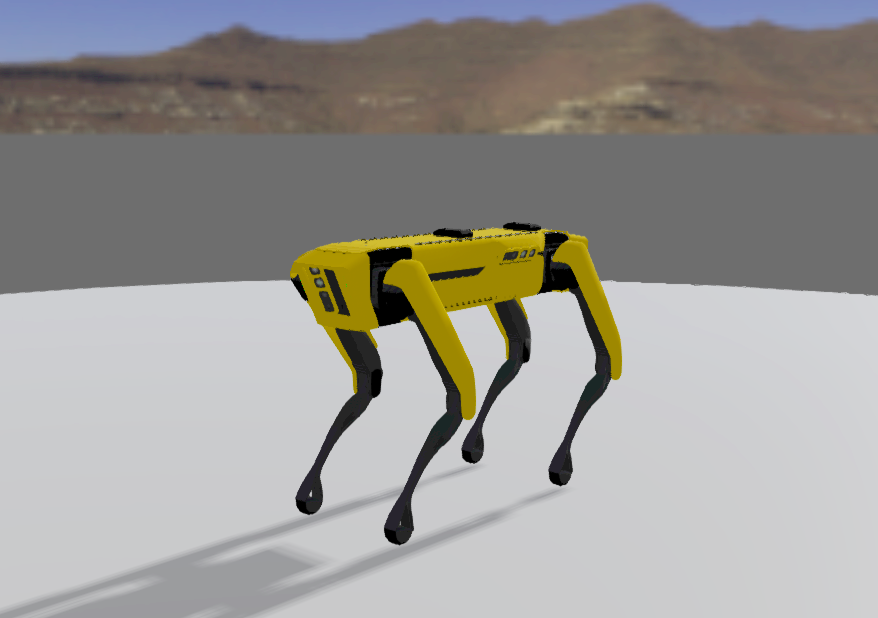}}
    \caption{Spot: Boston Dynamics robot in Webots}
    \label{fig:spot}
\end{figure}
In ~\cite{lit1}, Mahdi Khorram and Ali A. Moosavian presented the path planning of a robot on uneven terrain using LQR control for achieving stability. The paper describes a zero moment point in the 3D space, which consists of the robot height and height change when in motion. Thus, the controller reduces the change in the robot’s height and motion because the robot’s constant oscillatory motion may reduce stability. In ~\cite{lit2}, Jiaxin Guo $et$ $al$. proposed an MPC design for foot placement and planning of the Quadruped Robot. The Newton Euler formulation is used for modeling the system dynamics. The controller is then simulated with the linearised system on MATLAB and the results are noted. The controller was designed using the MPC toolbox from MATLAB which uses the system output feedback and the desired reference trajectory/path as the other input. In~\cite{mpccom}, Tomislav Horvat $et$ $al$. described the MPC for the quadruped robot locomotion. MPC is used to control and adjust the positions of the footsteps in order to satisfy stability constraints. MPC uses a desired centre of mass (CoM) trajectory projection along with the actual trajectory and works on matching the actual with the desired by reducing the error. In the case of Quadruped locomotion, where dynamic stability is a major concern while traversing on uneven terrains, MPC is the best suited as it predicts and changes the stability criteria according to the terrain ahead. In~\cite{nmpc} , Michael Neunert $et$ $al$. presented a nonlinear MPC approach to control the robot locomotion. A full-dynamic system model is used for experimentation and simulation. An Auto-Differentiable contact model used along with the dynamic system helps in efficiently optimizing the path planning. The model is tested for periodic gait patterns as well as highly dynamic motions like squat jumps. The proposed method shows satisfactory results on both robots.

In our work, instead of using the Inertial Measurement Unit (IMU) feedback to design the MPC, we use the ground reaction forces at each foot and design a torque based MPC. This concept was initially proposed in ~\cite{MIT} and ~\cite{ANYmal} and we build upon the same to develop torque based feedback for locomotion on inclined uneven terrains. In addition to this, we are also able to perform gait transitions, provide resistance to external perturbations and sustain falls using MPC. The torque based controller locks the joints at the desired values, thereby reducing vibrations and deviation from set-point. We used the readily available Spot robot by Boston Dynamics in Webots, fig.~\ref{fig:spot}, simulator for testing our controller. 

The rest of the paper is structured as follows: Section II talks about the leg kinematic equations for Spot. Section III briefs about the simplified dynamic model of the quadruped along with the state-space formulation of the robot. Section IV explains the torque based MPC in detail. The gait design and transitioning is explained in Section V. The MATLAB and Webots simulation results are presented in Section VI and the contributions are concluded in the final section.

\section{Kinematic Model}
The kinematic model design starts from the basic step of calculating the Denavit–Hartenberg (DH) parameters for each leg of the robot. The DH table helps understand the orientation of each link with respect to the previous one~\cite{kin}. Table~\ref{tab1: tab1} describes the DH parameters for Spot's leg configuration.
\begin{table}[htbp]
\caption{DH parameters for robot leg}
\begin{center}
\begin{tabular}{|c|c|c|c|c|}
\hline$(i-1)-i$ & $a_{i}$ & $\alpha_{i}$ & $d_{i}$ & $\theta_{i}$ \\
\hline $0-1$ & 0 & $90^{\circ}$ & $a_{1}$ & $\theta_{1}$ \\
\hline $1-2$ & $a_{2}$ & 0 & 0 & $\theta_{2}$ \\
\hline $2-3$ & $a_{3}$ & 0 & 0 & $\theta_{3}$ \\
\hline
\end{tabular}
\label{tab1: tab1}
\end{center}
\end{table}\\
In Table ~\ref{tab1: tab1}, $a$ is the length of the common normal, the distance between the $i^{th}$ and $i-1^{th}$ z-axis, $\alpha$ is the angle about the common normal between the $i-1^{th}$ z-axis and $i^{th}$ z-axis, $d$ is the distance between $i-1^{th}$ and $i^{th}$ x-axis, along $i-1^{th}$ z-axis and $\theta$ is the angle about the z-axis from the $i −1^{th}$ x-axis and the $i^{th}$ x-axis. $\theta_{1}$, $\theta_{2}$, $\theta_{3}$ are the hip roll, hip pitch, and the knee pitch angles of the leg. $a_{1}$ is the horizontal length between hip joint and the torso centre of mass,$a_{2}$ and $a_{3}$ are the lengths of upper leg and lower leg respectively. Using the above table, we develop the forward kinematics of the robot which are explained in the next section.
\subsection{Forward Kinematics}
 If the angular position of each leg joint is known at any point in time, then forward kinematics help find the coordinates of each joint of the leg in 3D space at that particular instance. The forward kinematics are mainly required to find the foot end-point coordinates using the joint angle values. The DH parameters are used to obtain the forward kinematic equations. Considering the abduction hip joint of each leg to be the origin, the forward kinematics for each leg in the body frame are calculated and are given in Table ~\ref{tab1: tab2}.\\
\begin{table}[htbp]
\caption{Quadruped Leg Forward Kinematics}
\begin{center}
\begin{tabular}{|c|c|c|c|c|}
\hline & $i=0$ & $i=1$ & $i=2$ & $i=3$ \\
\hline$x_{i}$ & 0 & 0 & $a_{2} C_{1} C_{2}$ & $C_{I} \left(a_{3} C_{12}+a_{2} C_{2}\right)$ \\
\hline$y_{i}$ & 0 & 0 & $a_{2} S_{l} C_{2}$ & $S_{1} \left(a_{3} C_{12}+a_{2} C_{2}\right)$ \\
\hline$z_{i}$ & 0 & $a_{1}$ & $a_{2} S_{2}+a_{l}$ & $a_{3} S_{23}+a_{2} S_{2}+a_{l}$ \\
\hline
\end{tabular}
\label{tab1: tab2}
\end{center}
\end{table}\\

where, $C_{i}=cos\theta_{i}$, $S_{i}=sin\theta_{i}$, $C_{ij}=cos(\theta_{i}+\theta_{j})$ and $S_{ij}=sin(\theta_{i}+\theta_{j})$

\subsection{Inverse Kinematics}
As the name explains, the inverse kinematics help find the joint angle values provided the end effector coordinates are known. So basically, the complete gait planning is based on the inverse kinematics of the quadruped legs. The step length and the semi-elliptical foot trajectory equation is programmed, and the joint angles at each instant are calculated using the inverse kinematic equations. The joint values are fed to the motors at the respective joints, resulting in the overall leg movement. The inverse kinematic model for the Quadruped leg is derived using the forward kinematic equations from Table ~\ref{tab1: tab2}.
\begin{equation}
    r=\sqrt{X_{3}^{2}+Y_{3}^{2}+Z_{3}^{2}+a_{1}^{2}-2a_{1}Z_{3}} 
\end{equation}
\begin{equation}
    \theta_{1}=tan^{-1}(\frac{Y_{3}}{X_{3}}) \\  
\end{equation}
\begin{equation}
    \theta_{2}=tan^{-1}(\frac{Z_{3}-a_{1}}{\sqrt{X_{3}^{2}+Y_{3}^{2}}})-cos^{-1}(\frac{a_{3}^{2}-a_{2}^{2}-r^{2}}{-2a_{2}r}) 
\end{equation}
\begin{equation}
    \theta_{3}=\pi-cos^{-1}(\frac{r^{2}-a_{2}^{2}-a_{3}^{2}}{-2a_{2}a_{3}}) 
\end{equation} \\ 
where, $X_{3}$, $Y_{3}$ and $Z_{3}$ are the foot coordinates in space. The next section briefs about the mathematical modeling of the quadruped dynamics. 
\begin{figure*}[!ht]
    \centering
    \centerline{\includegraphics[width = 488pt]{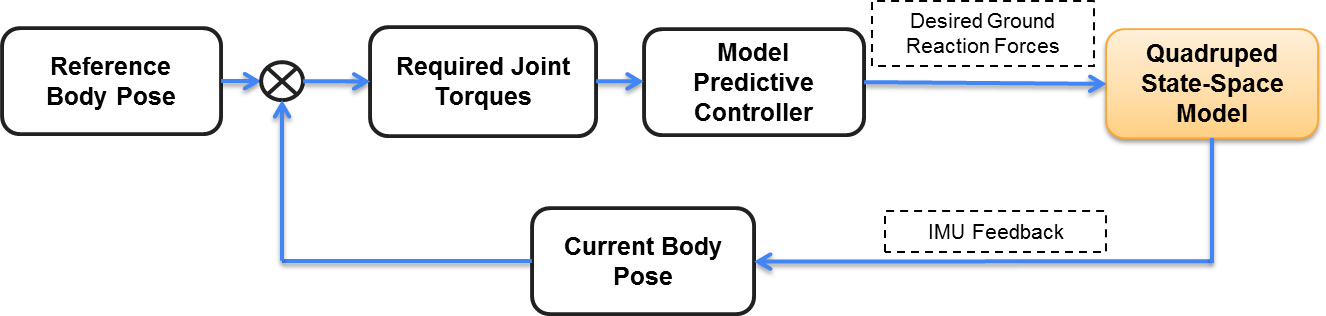}}
    \caption{The system block diagram}
    \label{fig:blockdiag}
\end{figure*}
\section{Quadruped Dynamics}
The dynamics, if derived using Lagrangian-Euler formulation, consist of complex non-linearity, which increases unnecessary computation cost and slow response. These equations can be simplified using appropriate approximations and modifying the non-linear system to a linear time-invariant system. We use the MIT mini Cheetah~\cite{MIT} approach for obtaining simplified quadruped dynamics. In this method, only the floating body, i.e., the torso orientation in the body frame with respect to the world frame, is considered. The leg dynamics are ignored, and only the distance between the foot and center of mass for each leg is considered in the state equation. This approximation can only be made when the total mass of the legs is less than 10\% of the total robot weight, which is true in the case of Spot. The foot endpoints are calculated using the kinematic equations. The state-space model of the robot is derived with A and B matrices, as mentioned above. The state parameters include the torso orientation in linear and angular coordinates along with their respective velocities. In addition to these 12 states, the gravity component forms the $13^{th}$ state. The input to the state-space model is the ground reaction forces on each foot in the linear space. The detailed state-space model is given in equation 6.

\begin{equation} \begin{bmatrix} \dot{\phi}\\ \dot{\theta}\\ \dot{\psi} \end{bmatrix} = \begin{bmatrix} \cos(\psi)\cos(\theta)&-\sin(\psi)/\cos(\theta)&0\\ -\sin(\psi)&\cos(\psi)&0\\ \cos(\psi)\tan(\theta)&\sin(\psi)\tan(\theta)&1 \end{bmatrix} {\omega} \end{equation}

\begin{align*}
 \frac{\mathrm{d}}{\mathrm{d}t} \begin{bmatrix} \hat{\Theta}\\ \hat{\mathrm{p}}\\ \hat{\omega}\\ \hat{\dot{\mathrm{p}}} \\ {\mathrm{g}} \end{bmatrix} = \begin{bmatrix} 0_{3\times3} & 0_{3\times3} & \mathrm{R}_{z}(\psi) & 0_{3\times3} & 0_{3\times1}\\ 0_{3\times3} & 0_{3\times3} & 0_{3\times3} & 1_{3\times3}  & 0_{3\times1}\\ 0_{3\times3} & 0_{3\times3} & 0_{3\times3} & 0_{3\times3} & 0_{3\times1}\\ 0_{3\times3} & 0_{3\times3} & 0_{3\times3} & 0_{3\times3} & G_{3\times1} \\0_{1\times3} & 0_{1\times3} & 0_{1\times3} & 0_{1\times3} & 1 \end{bmatrix} \begin{bmatrix} \hat{\Theta}\\ \hat{\mathrm{p}}\\ \hat{\omega}\\ \hat{\dot{\mathrm{p}}} \\ {\mathrm{g}} \end{bmatrix}
\end{align*}
\begin{equation}
\\ + \begin{bmatrix} 0_{3\times3} & 0_{3\times3} & 0_{3\times3} & 0_{3\times3}\\ 0_{3\times3} & 0_{3\times3} & 0_{3\times3} & 0_{3\times3}\\ \hat{\mathrm{I}}^{-1}[\mathrm{r}_{1}]_{\times} & \hat{\mathrm{I}}^{-1}[\mathrm{r}_{2}]_{\times} & 
\hat{\mathrm{I}}^{-1}[\mathrm{r}_{3}]_{\times} &
\hat{\mathrm{I}}^{-1}[\mathrm{r}_{4}]_{\times}\\ 
1_{3\times3}/m & 1_{3\times3}/m & 1_{3\times3}/m & 1_{3\times3}/m \\ 0_{1\times3} & 0_{1\times3} & 0_{1\times3} & 0_{1\times3} \end{bmatrix} \begin{bmatrix} \mathrm{f}_{1}\\ \vdots \\ \mathrm{f}_{12} \end{bmatrix} \end{equation}

\begin{equation} G_{3\times1} = \begin{bmatrix} 0 & 0 & 1 \end{bmatrix}^{T} \end{equation}
The state matrix from top to bottom, consists of joint angles [roll($\phi$), pitch($\theta$), yaw($\psi$)], cartesian position [x, y, z], joint velocities, linear velocities followed by the gravity component. ${R}_{z}$ is the rotation matrix of the body frame with respect to the world frame along z-axis. $0_{n\times m}$ is the null matrix and $1_{n\times n}$ is the identity matrix. $\hat{\mathrm{I}}$ is the inertia matrix in world frame and $[\mathrm{r}_{i}]_{\times}$ holds the distance between the center of mass and the foot endpoint. $m$ denotes the torso mass. Lastly, $\mathrm{f}_{i}$ denote the ground reaction forces on each foot. Once the dynamics are modelled, the MPC is designed. 

\section{Model Predictive Control}
As mentioned above, the input to the quadruped state-space model is the ground reaction forces experienced at each foot. Compared to the classical controllers, MPC has a few advantages, which proves it to be the superior one. First, MPC uses the system model, i.e., the robot dynamic, for predicting future steps based on the model's current state. This helps in estimating any error that may occur beforehand and correcting dynamically correcting it to avoid any repercussions. Second, the controller gains are designed based on a cost function for optimal performance and low overshoots. However, MPC considers the system constraints and limitations and takes the necessary action accordingly, which is not the case is classical optimal controllers like the Linear Quadratic Regulator (LQR). LQR shows better performance for unconstrained systems like Quadrotors~\cite{LQR} or Wheeled Robots. This makes MPC a better candidate for multi-constrained systems like legged robots. Apart from the system model and the tuned weights, the MPC needs the initial and the goal position. Depending on the update frequency, the controller estimates its state with respect to the goal position and recalculates its trajectory from that state towards the final position. This estimation and correction are possible since the controller itself uses the robot dynamics for its operation. 
\begin{figure*}[!ht]
    \centering
    \centerline{\includegraphics[width = 488pt]{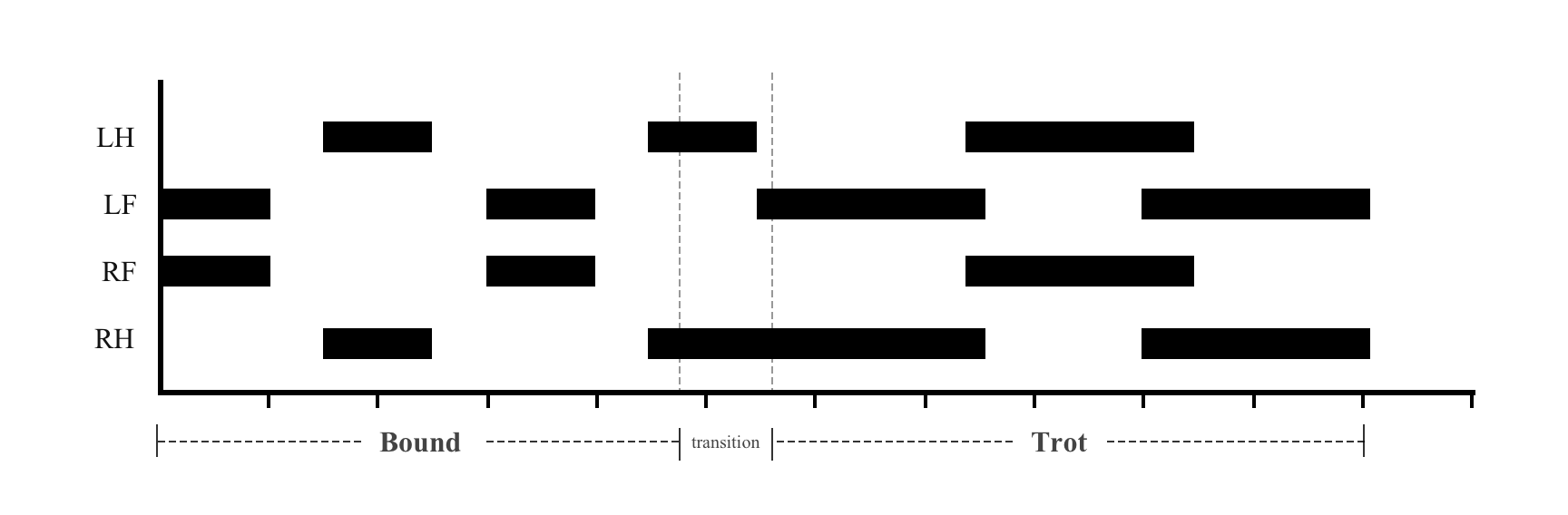}}
    \caption{Foot Trajectory from bound to trot}
    \label{fig:foottraj}
\end{figure*}
Here, the controller corrects the reaction forces, which indirectly means joint torque control of each quadruped leg. For a particular gait or a particular position, the joint torque values are fixed, which means that the ground reaction forces for that particular instance will also be fixed. Consider a simple standing position of the quadruped on flat ground. Since all four legs of Spot have the same kinematic structure, the jacobian matrix will be the same for all. On flat ground, all the 4 feet will experience equal ground reaction force in x, y, and z-direction. Such conditions for different scenarios are defined and the controller is designed. The complete system block diagram is given in fig.~\ref{fig:blockdiag}  

\section{Gait Definition}
There exist multiple quadrupedal gaits for varying speed and terrain definition. In this work, we have implemented the trot, which is meant for medium pace, and the bound gait for high speed locomotion. While bounding, the Left Front (LF) and Right Front (RF) legs move together followed by the Left Hind (LH) and the Right Hind (RH) legs. The phase difference between the Front and Hind legs is $180^{\circ}$. At a particular instance, only the front or the back legs are in the stance phase while the other two are in swing phase. 

For trot gait, the two diagonally opposite legs move together, i.e., [RH, LF] followed by [RF, LH]. This gait is meant for medium pace and is found to be more suitable for locomotion uneven terrains as well. The robot was made to execute the bounding gait on flat ground and the inclined uneven terrain in Webots. However, to our notice, the robot was not able to follow the terrain transition and got stuck. A feedback check is added to ensure the torso has a continuous motion. If there is no torso motion for defined period of time, the gait transitioning is executed from bound to trot. The gait cycle for bound to trot transition is shown in fig. ~\ref{fig:foottraj}. Gait transitioning is an unstable operation and may result in the robot losing balance and toppling~\cite{active}~\cite{HyQ}. It needs to be ensured that during the transition phase, the robots center of mass stays within the support polygon formed by the legs in stance phase. The MPC plays a crucial role in the transition phase\footnote{\url{https://youtu.be/FCjPmg-vUTg}}.      
The robot is able to successfully transition from bound to trot and is able to climb uneven terrains up to elevation of $20^{\circ}$.

\section{Results}
The non-linear state-space model of the quadruped robot is built in MATLAB, as explained in Section III. MATLAB provides an MPC toolbox that helps in designing the controller for the dynamic system. The controller requires the state-space model, a base reference model of the input and output signals, the input and output constraints, and the tuned weight matrices as per the optimal cost function. The MPC is designed using the toolbox, and the mathematical model of the system is simulated for various environmental conditions. The base reference model needs to be chosen appropriately to simulate the respective environmental conditions. For an uneven surface, a sine wave resembles correctly; for flat smooth ground, a constant signal will be preferred, whereas a ramp signal will rightly model an inclined surface. 
\begin{figure}[htbp]
    \centerline{\includegraphics[width = 244pt]{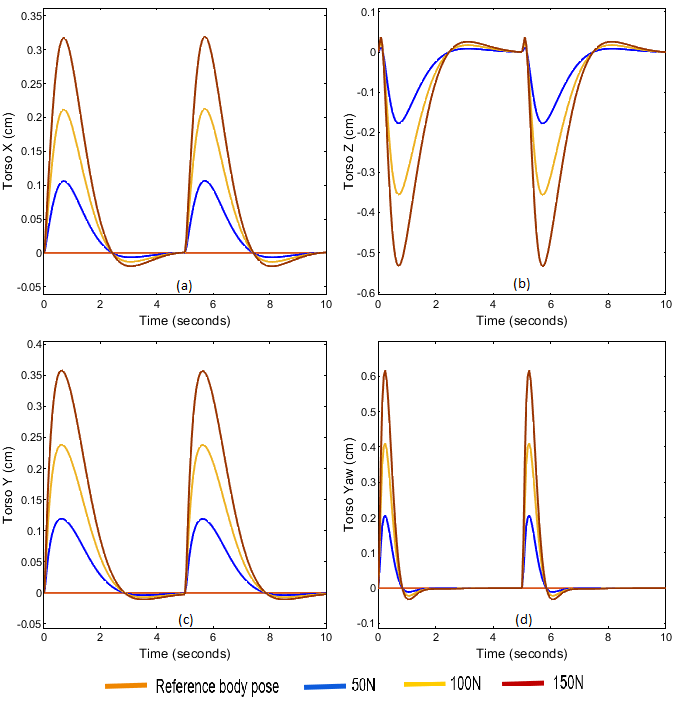}}
    \caption{Torso pose correction using MPC for increasing external disturbance magnitude}
    \label{fig:force}
\end{figure}
\begin{figure*}[!ht]
    \centering
    \centerline{\includegraphics[width = 488pt]{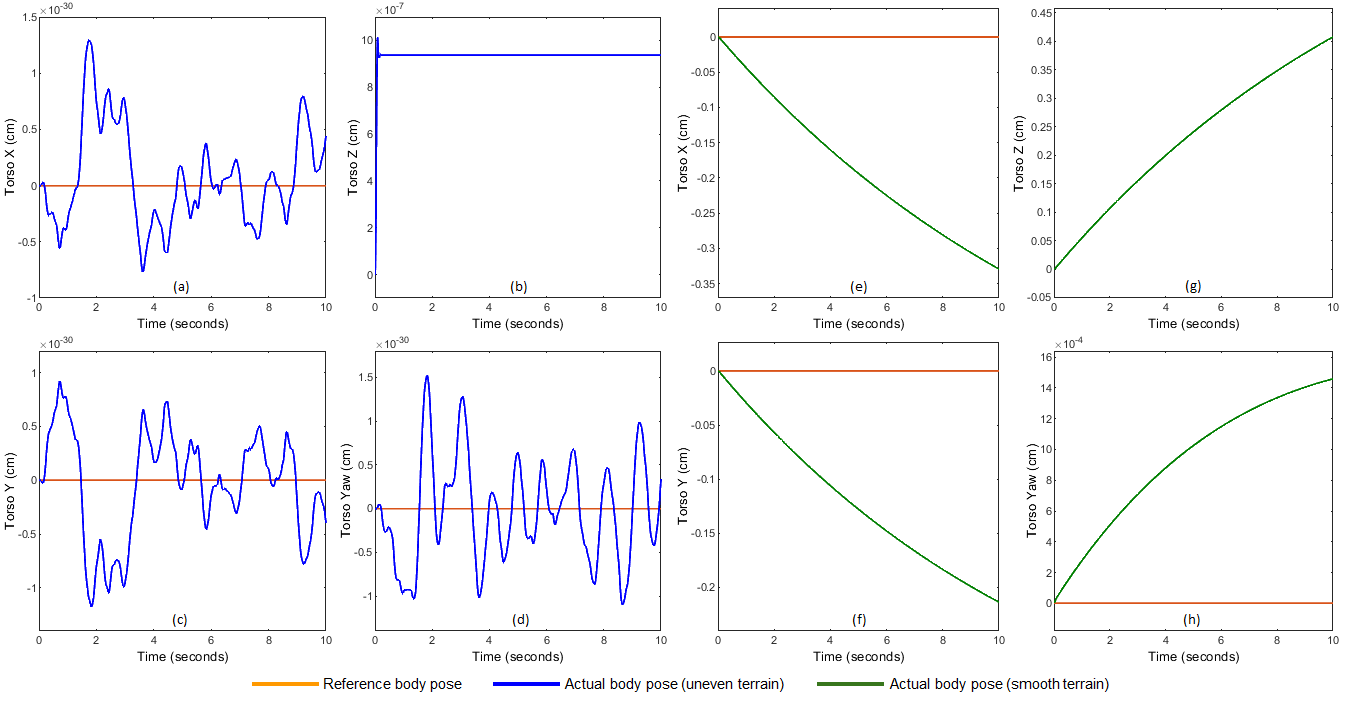}}
    \caption{Torso motion for uneven terrain for sine model (left) and constant model (right)}
    \label{fig:matlab}
\end{figure*}

This model is used as a reference for the end-effector trajectory while designing the MPC. It gives an estimate to the controller as to how the future state of the system might be so that it can rightly predict the action required to maintain stability. It must be noted that this model does not perfectly resemble the future system state but only helps in tuning the controller by assuming similar system nature. In case of high update frequency of MPC (50Hz and above), this reference model has negligible importance. This is because the controller estimates the system state and provides correction at a high rate. However, due to hardware limitations and high computation cost, it is not possible to run MPC at frequencies higher than 30 Hz on the actual robot. We have designed the controller for an update frequency of 20Hz with a sine reference model for simulating uneven terrain. At lower frequencies the reference model plays an important role which can be seen in fig~\ref{fig:matlab}. The graph represents the robot's motion on an uneven terrain recorded for 10 timesteps. For a stable motion, the torso orientation is desired to be [0, 0, z, 0] for [x, y, z, yaw] respectively. The blue curve (fig~\ref{fig:matlab} (a, b, c, d)) represents the actual torso motion when sine model was used and the green curve (fig~\ref{fig:matlab} (e, f, g, h)) represents the torso motion when the controller was designed using constant reference model. The torso stability is maintained at the set-point for the sine model whereas continuously increasing values are found in the constant model depicting unstable nature. In order to test the controller's performance for external disturbances, an impulse signal was given to the model with a time period of 5 s. The amplitude of the impulse signal was increased for finding the tolerable limit. The controller performance can be seen in fig.~\ref{fig:force}. The torso deviates when force is applied but controller corrects the body pose within 2 s time. The recovery time increases from 2 s to 4.5 s for 150 N disturbance which can be called as the limit. After this point, the next deviation will take place before initial correction which will increase instability and lead to toppling. The controller was tuned in MATLAB and tested on the mathematical model. Later the MPC was designed using gekko optimiser and validated in Webots\footnote{\url{https://youtu.be/5Sab5FzH6Qg}}.

\section*{Conclusion and Future Work}
The quadruped simplified dynamic equations are modelled and simulated in MATLAB. The MPC controller designed in MATLAB is tested for various environmental conditions and shows promising results. The controller is then verified for similar conditions in Webots. A smooth quadrupedal gait transition from high paced bounding to medium paced trot gait is executed successfully. MATLAB and Webots simulations show that the MPC can handle disturbances up to 150 N and is able to sustain falls from heights up to 80 cm. The robot is also able to traverse on inclined uneven surfaces with the help of the MPC.

The future work will focus on designing the other quadrupedal gaits and designing policies for the robot to achieve stable locomotion on more inclined surfaces. Environment mapping will be explored for dynamic trajectory optimisation on the run. We also plan on designing non-periodic gaits to allow locomotion through rocky and sandy terrain by predicting the safe foothold positions using Reinforcement Learning techniques.

\vspace{12pt}
\color{red}

\end{document}